\documentclass[journal]{IEEEtran}
\usepackage{cite}
\usepackage[dvipsnames]{xcolor}
\usepackage{graphicx}
\usepackage{amsmath}
\usepackage{amssymb,mathtools,commath}
\usepackage{algorithm,algorithmic}
\usepackage{array}
\usepackage{bbm}
\usepackage{stfloats}
\usepackage{url}
\usepackage[caption=false, font=footnotesize]{subfig}
\usepackage{tikz}
\usepackage{comment}
\usepackage{pgfplots}
\usepackage{multicol}
\usepackage{booktabs}
\usepackage{tabularx}
\usepackage[normalem]{ulem}
\usepackage{framed}
\usepackage{stackengine}
\usepackage{marginnote}
\usepackage{hyperref}

\DeclareMathOperator*{\argmin}{arg\,min}
\usepackage{soul}

\newcommand{\PT}{{\boldsymbol{\rho}}}

\newcommand{\AT}{\mathcal{A}}      
\newcommand{\vect}[1]{\mathbf{#1}}

\pgfplotsset{compat=newest}
\usetikzlibrary{positioning,spy,backgrounds,calc,arrows}
\pgfdeclarelayer{background}
\pgfdeclarelayer{foreground}
\pgfsetlayers{background,main,foreground}

\begin{document}
\title{MoDL-MUSSELS: Model-Based Deep Learning for Multishot Sensitivity-Encoded Diffusion MRI}

\author{Hemant~K.~Aggarwal,~\IEEEmembership{Member,~IEEE}, Merry~P.~Mani,
  and Mathews~Jacob,~\IEEEmembership{Senior~Member,~IEEE}
\thanks{Copyright (c) 2019 IEEE. Personal use of this material is permitted. However, permission to use this material for any other purposes must be obtained from the IEEE by sending a request to pubs-permissions@ieee.org.}
  \thanks{Hemant~K.~Aggarwal~(Email: hemantkumar-aggarwal@uiowa.edu) and Mathews Jacob~(Email: mathews-jacob@uiowa.edu) are with the department of electrical and computer engineering, University of Iowa, Iowa, USA.}
  \thanks{Merry Man~(Email: merry-mani@uiowa.edu) is with the division of neuroradiology, University of Iowa, Iowa, USA.}
        \thanks{This work is supported by 1R01EB019961‐01A1. This work was conducted on an MRI instrument funded by 1S10OD025025-01.  }
}

\maketitle
\begin{abstract}
  We introduce a model-based deep learning architecture termed MoDL-MUSSELS for the correction of phase errors in multishot diffusion-weighted echo-planar MR images.  The proposed algorithm is a generalization of the existing MUSSELS algorithm with similar performance but  significantly reduced computational complexity. In this work, we show that an iterative re-weighted least-squares implementation of MUSSELS alternates between a multichannel filter bank and the enforcement of data consistency. The multichannel filter bank projects the data to the signal subspace, thus exploiting the annihilation relations between shots.
  Due to the high computational complexity of the self-learned filter bank,  we propose replacing it with a convolutional neural network (CNN) whose parameters are learned from exemplary data.  The proposed CNN is a hybrid model involving a multichannel CNN in the k-space and another CNN in the image space. The k-space CNN exploits the annihilation relations between the shot images, while the image domain network is used to project the data to an image manifold. The experiments show that the proposed scheme can yield reconstructions that are comparable to state-of-the-art methods while offering several orders of magnitude reduction in run-time.
 
\end{abstract}

\IEEEpeerreviewmaketitle

\section{Introduction}

Diffusion MRI (DMRI), which is sensitive to anisotropic diffusion processes in the brain tissue, has the potential to provide rich
information on white matter anatomy\cite{le2003nature}. It has several applications, including the studies of neurological
disorders\cite{disorder2008lancet}, the aging process\cite{aging2006}, and acute stroke\cite{moseley1990early}. Diffusion MRI relies on large bipolar
directional gradients to encode water diffusion,  which attenuates the signals from diffusing molecules in the direction
of the gradient. The diffusion-sensitized signal is often spatially encoded using single-shot echo-planar imaging (ssEPI), which allows the acquisition of the entire k-space in a single excitation and readout. While such acquisitions can offer high sampling efficiency, the longer readout makes the acquisition vulnerable to distortions induced by B0 inhomogeneity. Specifically, the recovered images often exhibit geometric distortions \cite{wu2017image}. These artefacts, resulting from the long readouts, essentially limit the extent of k-space coverage and thereby the spatial resolution that ssEPI sequences can achieve.

\begin{figure}
  \centering \includegraphics[width=.6\linewidth]{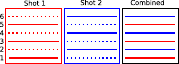}
  \caption{Demonstration of multishot EPI acquisition employing
    multiple excitations and readouts. The first radio-frequency~(RF) excitation and
    diffusion sensitization are followed by a k-space readout by shot 1
    that samples k-space lines 1, 3, and 5.  The second RF excitation
    and diffusion sensitization are followed by a k-space readout by
    shot 2 capturing lines 2, 4, and 6. The combined data corresponds to
    the fully sampled k-space.
  }
  \label{fig:msepi}
\end{figure}

Multishot echo-planar imaging~(msEPI) methods were introduced to minimize the distortions related to the long
readouts in ssEPI. This scheme splits the k-space sampling over multiple excitations and readouts, resulting in shorter readout lengths for each shot, as shown in Fig.~\ref{fig:msepi}. While multishot
imaging can offer high resolution, a challenge associated with this scheme is its vulnerability to inter-shot motion in the diffusion setting. Specifically, subtle physiological motion during the large bipolar gradients manifests as phase differences between different shots. The direct combination of the k-space data from these shots results in artefacts in the diffusion-weighted images (DWI) arising from phase inconsistencies. 

We recently introduced a multishot sensitivity-encoded diffusion data
recovery algorithm using structured low-rank matrix completion~(MUSSELS)~\cite{mussels}, which allows the  reconstruction of DWI that are immune to the motion-induced phase artefacts. The method exploits the redundancy between the Fourier samples of the shots to jointly
recover the missing k-space samples in each of the shots~\cite{irlsMussels}. The k-space data recovery is then posed as a matrix completion problem that utilizes a structured low-rank algorithm and parallel imaging to recover the missing k-space data in each shot.  While this scheme can offer state-of-the-art results, the challenge is the high computational complexity.
The large data size and the need for matrix lifting make it challenging to reconstruct the high-resolution data from different directions and slices despite the existence of fast structured low-rank algorithms~\cite{irlsMussels,gregGIRAF2017}.

In this paper, we introduce a novel deep learning framework to minimize the computational complexity of MUSSELS~\cite{mussels}. This work is inspired by the convolutional network structure of MUSSELS and is  formulated in k-space to exploit the convolutional relations between
the Fourier samples of the shots. The proposed scheme is also motivated by our recent work on model-based deep learning (MoDL) \cite{modl} and similar algorithms that rely on the unrolling of iterative algorithms \cite{adler2018learned,admmnet,hammernik}. The main benefit of MoDL is the ability to exploit the physics of the acquisition scheme and the ability to incorporate multiple regularization priors \cite{modl-storm}, in a deep learning setting, to achieve improved performance.
The use of the conjugate-gradient algorithm within the network to enforce data consistency in MoDL provides improved performance for a specified number of iterations.
The sharing of network parameters across iterations enables MoDL to keep the number
of learned parameters decoupled from the number of iterations, thus providing good convergence without increasing the number of trainable parameters.  A smaller number of trainable parameters translates to significantly reduced training data demands, which is particularly attractive for  data-scarce  medical-imaging applications.

We first introduce an approach based on the iterative reweighted least-squares algorithm (IRLS)~\cite{irls}  to solve the MUSSELS cost function~\cite{mussels}. The MUSSELS algorithm~\cite{mussels}, which is based on iterative singular value shrinkage, alternates between a data-consistency block and a low-rank matrix recovery block. By contrast, the IRLS-MUSSELS algorithm~\cite{irlsMussels} alternates between a data-consistency block and a residual  multichannel\footnote{The term multichannel is used in a traditional signal processing sense to refer to multichannel convolution using a multichannel filter.  The different channels corresponds to the images from different shots of the multishot data.  Since we rely on the SENSE forward model~\cite{sense1999}, the channels do not refer to multi-coil data in parallel MRI. } convolution block. The multichannel convolution block can be viewed as the projection of the data to the nullspace of the multichannel signals; the subtraction of the result from the original ones, induced by the residual structure, projects the data to the signal subspace, thus removing the artefacts in the signal.
The IRLS-MUSSELS algorithm learns the parameters of the denoising filter from the data itself, which requires several iterations. Motivated by our earlier work~\cite{modl}, we propose replacing the multichannel linear convolution block in IRLS-MUSSELS with a convolutional neural network (CNN). Unlike the self-learning strategy in IRLS-MUSSELS, where the filter parameters are learned from the measured data itself, we propose learning the parameters of the non-linear CNN from exemplar data. We hypothesize that the non-linear structure of the CNN will enable us to learn and generalize from examples. The learned CNN will facilitate the projection of each test dataset to the associated signal subspace. While the architecture is conceptually similar to MoDL, the main difference is the extension to multishot settings and the learning in the Fourier domain (k-space) enabled by the IRLS-MUSSELS reformulation.

The proposed framework has similarities to recent k-space deep learning strategies \cite{jongEPIghost,wnet,wang2018dimension,kikinetMRM}, which also exploit the convolution relations in the Fourier domain. The main
distinction of the proposed scheme with these methods is the model-based framework, along with the training of the unrolled
network. Many of the current schemes \cite{wang2018dimension} are not designed for the parallel imaging setting. The use of the conjugate gradient steps in our network allows us to account for parallel imaging efficiently, requiring few iterations. We also note the relation of the proposed work with that of Akcakaya et al.~\cite{mehmet2018}, which uses a self-learned network to recover parallel MRI data.  The weights of the network are estimated from the measured data itself. Since we estimate the weights from exemplar data, the proposed scheme is significantly faster.

\section{Background}
\subsection{Problem formulation}
The high-resolution DMRI requires long-duration EPI readouts that are vulnerable to field-inhomogeneity-induced spatial distortions.
Also, the large rewinder gradients make the achievable echo-time rather long, resulting in lower signal-to-noise ratio~(SNR). To minimize these distortions, it is common practice to acquire the data using msEPI schemes for high-resolution applications. These schemes acquire a highly undersampled subset of the k-space at each shot. Since the subsets are complementary, the data from all these shots can be combined to obtain the final image. The image acquisition of the $i^{\rm th}$ shot and the $j^{\rm th}$ coil can be expressed as 
\begin{equation}\label{fwd}
 y_{i,j}[\mathbf k] = \int_{\mathbb R^2} \rho(\mathbf r) s_j(\mathbf r) \exp\left(\mathbbm{i}~\mathbf k^T \mathbf r\right) d\mathbf r +  n_{i,j}[\mathbf k];~\forall \mathbf k \in \Theta_i.
\end{equation}
Here, $\mathbf s_j(\mathbf r)$ denotes the coil sensitivity of the $j^{\rm th}$ coil and $\Theta_i, i=1,..,N$, denotes the subset of the k-space that is acquired at the $i^{\rm th}$-shot. Note that the sampling indices of the different shots are complementary, implying that the combination of the data from the different shots will result in a fully sampled image. Specifically, we have $\bigcup_{i=1}^{N} \Theta_i = \Theta$, where $\Theta$ is the Fourier grid corresponding to the fully sampled image. The above relation, to acquire the desired image $\rho(\mathbf r)$ from $N$ shots, can be compactly represented as 
\begin{equation}
  \label{eq:ImageFormationModel}
\mathbf y_i = \mathcal{A}_i({\rho}(\mathbf r)) + \mathbf n, ~~ i=1,..,N
\end{equation}
in the absence of phase errors. Here, $\mathbf y_i$ represents the undersampled multichannel measurements of the $i^{\rm th}$ shot acquired using the acquisition operator $\mathcal A_i$, and $\mathbf n$ represents the additive Gaussian noise that may corrupt the samples during acquisition. 

Diffusion MRI uses large bipolar diffusion gradients to encode the diffusion motion of water molecules. Unfortunately, subtle physiological motion between the bipolar gradients often manifests as phase errors in the acquisition. With the addition of the unknown phase function $\phi_i(\mathbf r), |\phi_i(\mathbf r)|=1$ introduced by physiological motion, the forward model is  modified as 
\begin{equation}
\label{modifiedforwardmodel}
\mathbf y_i = \mathcal{A}_i\Big (\underbrace{{\rho}~(\mathbf r) \phi_i(\mathbf r)}_{\rho_i(\mathbf r)}\Big ) + \mathbf n, ~~ i=1,..,N.
\end{equation}
If the phase errors $\phi_i(\mathbf r), i=1,..,N$, are uncompensated, the image obtained by the combination of $\mathbf y_i,\, i=1,..,N$ will show artefacts arising from the inconsistent phase. {The widely used multishot method, termed  MUSE~\cite{muse,pocsmuse}, relies on the independent estimation of $\phi_i(\mathbf r)$ from low-resolution reconstructions of the phase-corrupted images $\rho_i(\mathbf r)$. The forward model can be compactly written as $\mathbf y = \mathcal A(\boldsymbol{\rho})$, where $
{\boldsymbol{\rho}}  = \begin{bmatrix}
{\boldsymbol{\rho}_1}^T, &
\ldots&
{\boldsymbol{\rho}_N}^T 
\end{bmatrix}^T
$ is the vector of multishot images. Once the phases are estimated, the reconstruction is posed as a phase-aware reconstruction \cite{muse,pocsmuse}.

\subsection{Brief Review of MUSSELS}
The  MUSSELS algorithm~\cite{mussels} relies on a structured low-rank formulation to jointly recover the phase-corrupted images $\rho_i$ from their under-sampled multi-coil measurements. The MUSSELS algorithm capitalizes on the  multi-coil nature of the measurements as well as annihilation relations between the phase-corrupted images. The key observation is that these phase-corrupted images satisfy an image domain annihilation relation~\cite{mathewsISBI2007}
\begin{equation}
\label{eq:imgAnnihilation}
\rho_i(\mathbf r)\phi_j(\mathbf r)-\rho_j(\mathbf r)\phi_i(\mathbf r) =0,\,  \forall \mathbf r.
\end{equation}
 This multiplicative  annihilation relation, resulting from phase inconsistencies, translates to convolution relations in the Fourier domain:
\begin{equation}\label{convrelations}
\widehat{\rho_i}(\mathbf k)\ast\widehat{\phi_j}(\mathbf k)-\widehat{\rho_j}(\mathbf k)\ast\widehat{\phi_i}(\mathbf k) =0 \quad \forall \mathbf k,
\end{equation}
where $\widehat x$ denotes the Fourier transform of $x$. Since the phase images $\phi_j(\mathbf r)$ are smooth, their Fourier coefficients $\widehat{\phi_j}(\mathbf k)$ can be assumed to be support-limited to a region $\Lambda$ in the Fourier domain. This allows us to rewrite the convolution relations in \eqref{convrelations} in a matrix form using block-Hankel  convolution matrices $\mathbf H_{\Lambda}^{\Gamma}(\rho)$. The matrix product $\mathbf H_{\Lambda}^{\Gamma}(\rho)~\mathbf s$ corresponds to the~2D convolution between a signal $\rho$ supported on a grid $\Gamma$ and the filter $\mathbf s$ of size $\Lambda$. Thus, the Fourier domain convolution relations can be compactly expressed using  matrices~\cite{mussels} as
\begin{equation}\label{eq:convInMatrix}
  \left[\mathbf H_{\Lambda}^{\Gamma}(\hat \rho_i) |\mathbf H_{\Lambda}^{\Gamma}(\hat \rho_j)\right]
  \underbrace{\begin{bmatrix}
 \hat{\boldsymbol{\phi_j}}\\
\overline{ -\hat{\boldsymbol{\phi_i}}}
 \end{bmatrix} }_{\mathbf s} = \mathbf 0.
\end{equation}
We note that there exists a similar annihilation relation between each pair of shots, which implies that the structured matrix 
\begin{equation}\label{block}
\mathbf{T}(\widehat{\boldsymbol{\rho}}) = \begin{bmatrix}  \mathbf H_{\Lambda}^{\Gamma}(\hat \rho_1) ~|~ \cdots ~|~ \mathbf H_{\Lambda}^{\Gamma}(\hat \rho_{N})  \end{bmatrix}
\end{equation}
is low-rank.  The MUSSELS algorithm\cite{mussels} recovers the multishot images  from their undersampled k-space measurements by solving
\begin{equation}
  \label{eq:mu}
  \tilde{{\boldsymbol{\rho}} }=\argmin_{{\boldsymbol{\rho}} } \norm{\mathcal {A}({\boldsymbol{\rho}}  )-\mathbf y }_2^2  + \lambda \norm{\mathbf{T}\left(\widehat{\boldsymbol{\rho}} \right)}_*,
\end{equation}
where $\|\cdot\|_*$ denotes the nuclear norm. 
The above problem is solved in earlier work~\cite{mussels} using an iterative shrinkage algorithm.

\section{Deep learned MUSSELS}
\subsection{IRLS reformulation of MUSSELS}

To bring the MUSSELS framework to the MoDL setting, we first introduce
an IRLS reformulation \cite{irls} of the MUSSELS. Using an auxiliary variable $\mathbf z$, we rewrite
\eqref{eq:mu} as
\begin{equation}
  \label{eq:mu1}
  \argmin_{\boldsymbol{\rho},\mathbf z} ~\norm{\mathcal {A}(\boldsymbol{\rho})-\mathbf y }_2^2  + \beta\|\widehat{\boldsymbol{\rho}}-\mathbf z\|_F^2 + \lambda \| \mathbf{T}({\mathbf z}) \|_*.
\end{equation}
We observe that \eqref{eq:mu1} is equivalent to \eqref{eq:mu} as
$\beta\rightarrow \infty$. An alternating minimization algorithm to
solve the above problem yields the following steps:
\begin{eqnarray}
  \label{dcMussels}
  \boldsymbol{\rho}_{n+1}&=&\argmin_{\boldsymbol{\rho}} \norm{\mathcal {A}(\boldsymbol{\rho} )-\mathbf y }_2^2  + \beta~\|\widehat{\boldsymbol{\rho}}-\mathbf z_n\|_F^2 \\\label{zproblem}
  \mathbf z_{n+1}&=&\argmin_{\mathbf z} \|\widehat{\boldsymbol{\rho}}_{n+1}-\mathbf z\|_F^2 + \frac{\lambda}\beta ~\| \mathbf{T} ({\mathbf z}) \|_*.
\end{eqnarray}
We now borrow from the literature~\cite{mohan2012iterative,fornasier2011low} and majorize the
nuclear norm term in \eqref{zproblem} as 
\begin{equation}\label{majorize}
  \norm{\mathbf{T} (\mathbf z)}_*\leq  \norm{ \mathbf{T} (\mathbf z) \mathbf Q}_F^2,
\end{equation}
where the weight matrix is specified by
\begin{equation}\label{weightupdate}
  \mathbf Q = \big[
  \mathbf{T}^H(\mathbf z)\mathbf{T}(\mathbf z) + \epsilon 
  \mathbf I\big] ^{-1/4}
\end{equation} 
Here, $\mathbf I$ is the identity matrix. Similar majorization strategies were used in the work~\cite{gregGIRAF2017}. With the majorization in~\eqref{majorize}, the $\mathbf z$-subproblem in~\eqref{zproblem} would involve the
alternation between
\begin{equation}
  \label{denoise}
  \mathbf z_{n+1}=\argmin_{\mathbf z} \|\widehat{\boldsymbol{\rho}}_{n+1}-\mathbf z\|_F^2 + \frac{\lambda}\beta ~\| \mathbf{T} ({\mathbf z})\mathbf Q \|_F^2
\end{equation}
and the update of the $\mathbf Q$ using~\eqref{weightupdate}.  Thus the IRLS reformulation of the MUSSELS scheme would alternate between~\eqref{dcMussels},~\eqref{denoise}, and~\eqref{weightupdate} as summarized in Algorithm~\ref{algo:irls}. The matrix $\mathbf Q$ may be viewed as a
surrogate for the nullspace of $\mathbf T(\mathbf z)$ as shown in the work~\cite{gregGIRAF2017}. The $\mathbf Q$ matrix at each iteration is estimated based on the previous iterate of $\mathbf z$.
The update
step~\eqref{denoise} can be interpreted  as finding an approximation of
$\widehat{\boldsymbol{\rho}}_{n+1}$ from the signal subspace.
\begin{algorithm}
  \caption{Summary of the IRLS-MUSSELS algorithm}
  \label{algo:irls}
 \begin{algorithmic}[1]
 \renewcommand{\algorithmicrequire}{\textbf{Input:}}
 \renewcommand{\algorithmicensure}{\textbf{Output:}}
 \REQUIRE  $\rho_0$, $z_0$
 \ENSURE  $\rho_{n+1}$
 \FOR {$n = 1$ to $\text{max\_Iterations}$}
  \STATE $\rho_{n+1} =$ solve~\eqref{dcMussels} using conjugate gradient
 \STATE $z_{n+1} =$ solve \eqref{denoise} using conjugate gradient.
 \STATE $ \mathbf Q_{n+1} = \big[
  \mathbf{T}^H(\mathbf z_{n+1})\mathbf{T}(\mathbf z_{n+1}) + \epsilon 
  \mathbf I\big] ^{-1/4}$
  \ENDFOR
 \RETURN $\rho_{n+1}$ 
 \end{algorithmic} 
 \end{algorithm}

\subsection{Interpretation of IRLS- MUSSELS as an iterative denoiser}

We now focus on the term $\|\mathbf{T} ({\mathbf z})\mathbf Q\|^2 = \sum_i \|\mathbf{T} ({\mathbf z})\mathbf q_i\|^2$ in \eqref{denoise}; $\mathbf q_i$ are the columns of $\mathbf Q$ representing nullspace vectors of $\mathbf{T} ({\mathbf z})$ similar vector $\mathbf s$ in  Eq.~\eqref{eq:convInMatrix}. We note that the matrix-vector product $\mathbf T(\mathbf z)\mathbf q_i$ corresponds to the multichannel convolution of $\mathbf z$ with the columns of $\mathbf Q$, specified by $\mathbf q_i$. We split each column $\mathbf q_i$ into sub-filters $\mathbf q_{ij}$ to obtain
\begin{equation}\label{key}
  \mathbf Q = \begin{bmatrix}
  \underbrace{\begin{bmatrix}
   \underline{\mathbf q_{11} }\\ \vdots \\ \overline{\mathbf q_{1N}}
  \end{bmatrix}}_{\mathbf q_1}&
 \underbrace{\begin{bmatrix}
	 \underline{\mathbf q_{21}}\\ \vdots \\ \overline{\mathbf q_{2N}}
	\end{bmatrix}}_{\mathbf q_2}&&
 \underbrace{\begin{bmatrix}
	\underline{\mathbf q_{N1}}\\ \vdots\\ \overline{\mathbf q_{NN}}
	\end{bmatrix}}_{\mathbf q_N}
  \end{bmatrix}
\end{equation}
where each $ \mathbf q_{ij}$ is of length $|\Lambda|$. Note that $\mathbf z = \begin{bmatrix}
\mathbf z_1\vline&\dots&\vline\mathbf z_N
\end{bmatrix}$ is the multishot data. This allows us to rewrite the multichannel convolution
\begin{equation}\label{expand}
	\mathbf T(\mathbf z) \mathbf q_i = \mathbf
	H_{\Lambda}^{\Gamma}(\mathbf z_1) \mathbf q_{i1} + ..\mathbf
	H_{\Lambda}^{\Gamma}(\mathbf z_N) \mathbf q_{iN}....
\end{equation}
as the sum of convolutions of $\mathbf z_j$ with $\mathbf q_{i,j}$. Due to the commutativity of convolution $h*g=g*h$, each term in \eqref{expand} can be re-expressed as 
\begin{equation}\label{commutative}
\mathbf H_{\Lambda}^{\Gamma}(\mathbf g)\mathbf h = \mathbf S(\mathbf h)\mathbf g,
\end{equation}
where $\mathbf S(\mathbf h)$ is an appropriately\footnote{The size of the matrix is $|\Gamma|-|\Gamma\ominus \Lambda|\times |\Gamma|$ such that \eqref{commutative} holds. Here, $\Gamma$ is the size of the image, and $\Lambda$ is the size of the filter. $\ominus$ refers to the set erosion operator as defined in the work~\cite{gregGIRAF2017}.} sized block Hankel matrix constructed from the zero-filled entries of $\mathbf h$.
We use this relation to rewrite 

\begin{equation*}
  \mathbf{T}\left({{\mathbf z}}\right)\mathbf Q=\underbrace{\begin{bmatrix}
      \mathbf S(\mathbf q_{11}) & \mathbf S(\mathbf q_{12})&\ldots & \mathbf S(\mathbf q_{1N})\\
      \vdots&\ldots&&\vdots\\
      \mathbf S(\mathbf q_{N1}) & \mathbf S(\mathbf q_{12})&\ldots & \mathbf S(\mathbf q_{NN})\\
    \end{bmatrix}}_{\mathbf{G}\left({\mathbf
        Q}\right)}\underbrace{\begin{bmatrix}
      \mathbf z_1\\
      \vdots\\
      \mathbf z_N
    \end{bmatrix}}_{\mathbf z }.
\end{equation*}
We note that $\mathbf G(\mathbf Q)\mathbf z$ corresponds
to the multichannel convolution of $\mathbf z_1,\dots,\mathbf z_N$ with the filterbank having filters $\mathbf q_{i,j}$. With this reformulation, \eqref{denoise} is simplified as
\begin{equation}
  \label{eq:irls}
  \mathbf z_{n+1}=\argmin_{\mathbf z}\|\widehat{\boldsymbol{\rho}}_n-\mathbf z\|_F^2 +  \frac\lambda\beta \left\|\mathbf{G}\left(\mathbf Q\right)\mathbf z\right\|_F^2.
\end{equation}
\begin{figure}	
  \centering
  \subfloat[Representation of  Eq.~\eqref{approxsoln} as the IRLS-MUSSELS denoiser $\mathcal D_w$.]{ \centering
    \def\svgwidth{.9\linewidth} 
\begingroup%
  \makeatletter%
  \providecommand\color[2][]{%
    \errmessage{(Inkscape) Color is used for the text in Inkscape, but the package 'color.sty' is not loaded}%
    \renewcommand\color[2][]{}%
  }%
  \providecommand\transparent[1]{%
    \errmessage{(Inkscape) Transparency is used (non-zero) for the text in Inkscape, but the package 'transparent.sty' is not loaded}%
    \renewcommand\transparent[1]{}%
  }%
  \providecommand\rotatebox[2]{#2}%
  \newcommand*\fsize{\dimexpr\f@size pt\relax}%
  \newcommand*\lineheight[1]{\fontsize{\fsize}{#1\fsize}\selectfont}%
  \ifx\svgwidth\undefined%
    \setlength{\unitlength}{153.74306126bp}%
    \ifx\svgscale\undefined%
      \relax%
    \else%
      \setlength{\unitlength}{\unitlength * \real{\svgscale}}%
    \fi%
  \else%
    \setlength{\unitlength}{\svgwidth}%
  \fi%
  \global\let\svgwidth\undefined%
  \global\let\svgscale\undefined%
  \makeatother%
  \begin{picture}(1,0.24391322)%
    \lineheight{1}%
    \setlength\tabcolsep{0pt}%
    \put(0,0){\includegraphics[width=\unitlength,page=1]{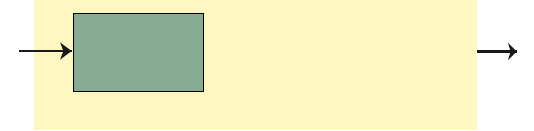}}%
    \put(0.20144009,0.1588649){\color[rgb]{0,0,0}\makebox(0,0)[lt]{\lineheight{1.25}\smash{\begin{tabular}[t]{l}Conv\end{tabular}}}}%
    \put(0,0){\includegraphics[width=\unitlength,page=2]{dwMuss.pdf}}%
    \put(0.48983718,0.15880356){\color[rgb]{0,0,0}\makebox(0,0)[lt]{\lineheight{1.25}\smash{\begin{tabular}[t]{l}DeConv\end{tabular}}}}%
    \put(0,0){\includegraphics[width=\unitlength,page=3]{dwMuss.pdf}}%
    \put(-0.03559642,0.13083744){\color[rgb]{0,0,0}\makebox(0,0)[lt]{\lineheight{1.25}\smash{\begin{tabular}[t]{l}$\widehat{\rho_n}$\end{tabular}}}}%
    \put(0.98395904,0.13005703){\color[rgb]{0,0,0}\makebox(0,0)[lt]{\lineheight{1.25}\smash{\begin{tabular}[t]{l}$\widehat{z_n}$\end{tabular}}}}%
    \put(0.20950213,0.09991036){\color[rgb]{0,0,0}\makebox(0,0)[lt]{\lineheight{1.25}\smash{\begin{tabular}[t]{l}$G(Q)$\end{tabular}}}}%
    \put(0.51317454,0.09747147){\color[rgb]{0,0,0}\makebox(0,0)[lt]{\lineheight{1.25}\smash{\begin{tabular}[t]{l}$G(Q)^H$\end{tabular}}}}%
  \end{picture}%
\endgroup%


    \label{subfig:dwMU}
  }
	
  \subfloat[The IRLS-MUSSELS algorithm ]{ \centering
    \def\svgwidth{\linewidth} \input{mussels.tex}
    \label{subfig:dwdcMU}
  }
	
  \caption{ (a). The interpretation of Eq.~\eqref{approxsoln} as a
    convolution-deconvolution network.  (b)  The IRLS-MUSSELS iterates
    between~\eqref{approxsoln} and~\eqref{dcMussels}. The data consistency (DC) step represents the solution of Eq.~\eqref{dcMussels}.}
  \label{fig:mussels}
\end{figure}

Differentiating the above expression and setting it equal to zero, we get
\begin{equation*}
  \mathbf z_{n+1}=  \left(\mathbf I\,+ \,\frac{\lambda}{\beta}~ \mathbf{G}\left(\mathbf Q\right)^H\mathbf{G}\left(\mathbf Q\right)\right)^{-1}\boldsymbol{\rho}_{n+1}.
\end{equation*}
One may use a numerical solver to determine $\mathbf z_{n+1}$. An
alternative is to solve this step approximately using the matrix
inversion lemma, assuming $\lambda << \beta$:
\begin{eqnarray}\nonumber
  \mathbf z_{n+1} &\approx&   \left[\mathbf I- \frac{\lambda}{\beta}\, \mathbf{G}\left(\mathbf Q\right)^H\mathbf{G}\left(\mathbf Q\right)\right]~\widehat{\boldsymbol{\rho}}_{n+1}\\\label{approxsoln}
                  &=&\widehat{\boldsymbol{\rho}}_{n+1} -  \frac{\lambda}{\beta}\, \mathbf{G}\left(\mathbf Q\right)^H\mathbf{G}\left(\mathbf Q\right)~\widehat{\boldsymbol{\rho}}_{n+1}.
\end{eqnarray}
We note that $G(\mathbf Q)$ can be viewed as a single layer
convolutional filter bank, while multiplication by $G(\mathbf Q)^H$
can be viewed as  flipped convolutions
(deconvolutions in deep learning context) with matching boundary
conditions. Note that neither of the above layers have any
non-linearities. Thus, \eqref{approxsoln} can be thought of as a residual
block, which involves the convolution of the multishot signals
$\widehat{\boldsymbol{\rho}}_n$ with the columns of $\mathbf Q$, followed by
deconvolution as shown in Fig.~\ref{subfig:dwMU}. As discussed before,
the filters specified by the columns of $\mathbf Q$ are surrogates for
the nullspace of $\mathbf T(\widehat{\boldsymbol{\rho}})$. Thus, the update \eqref{approxsoln} can be thought of as removing the components of
$\widehat{\boldsymbol{\rho}}_n$ in the nullspace and projecting the data to
the signal subspace, which may be viewed as a \emph{sophisticated
  denoiser}, as shown in Fig.~\ref{subfig:dwMU}. 

The IRLS-MUSSELS scheme~\cite{irlsMussels}, as summarized in Fig.~\ref{fig:mussels}, provides state-of-the-art results. However, note that the filters specified by the
columns of $\mathbf Q$ are estimated for each diffusion direction by using Algorithm~\ref{algo:irls}, which has high computational complexity, especially in the context of
diffusion-weighted imaging, where several directions need to be
estimated for each slice.

\subsection{MoDL-MUSSELS Formulation}
\label{sec:kspace}

\begin{figure}
  \centering 
  \subfloat[The M-layer CNN-based denoiser \label{subfig:cnn}]{
    \def\svgwidth{.93\linewidth}
    \input{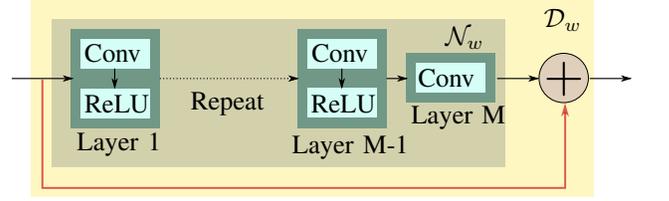} }

  \subfloat[Proposed k-space MoDL-MUSSELS architecture \label{subfig:dwdcKsp}]{
    \def\svgwidth{.99\linewidth} \input{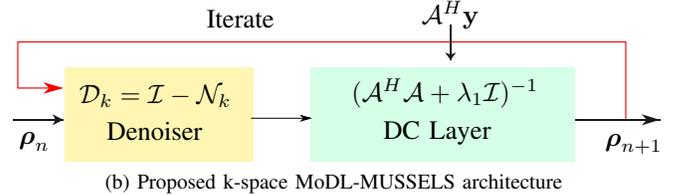} }

  \caption{The block diagram of the proposed k-space network architecture to solve Eq.~\eqref{eq:modlKsp}.  (a) The $\mathcal N_w$ block represents the deep learned noise predictor, and $\mathcal D_w$ is a residual learning block. (b) Here, the denoiser $D_k$  is  the M-layer network $\mathcal D_w$ that performs  the k-space denoising. }
  \label{fig:kspaceModel}
\end{figure}

To minimize the computational complexity of the IRLS-MUSSELS, we propose learning a non-linear \emph{denoiser} from exemplar data rather than
learning a custom denoising block specified by
$ \left[\mathbf I- \frac{\lambda}{\beta}\, \mathbf{G}\left(\mathbf
    Q\right)^H\mathbf{G}\left(\mathbf Q\right)\right]$ for each
direction and slice. We hypothesize that the non-linearities in the
network, as well as the larger number of filter layers, can facilitate
the learning of a generalizable model from the exemplar data. This
framework may be viewed as a multishot extension of the MoDL~\cite{modl} approach. The cost function associated with the network is
\begin{equation}
  \label{eq:modlKsp}
  \argmin_{{ \boldsymbol{\rho}}} \|\mathcal {A}({ \boldsymbol\rho} )-\mathbf y \|_2^2  + \lambda_1 \|\mathcal N_k(\boldsymbol{ \rho})\|_2^2. 
\end{equation}

Here, $\mathcal N_k(\boldsymbol{\rho})$ is a non-linear residual
convolutional filterbank working in the Fourier domain, with
\begin{equation}\label{residual}
  \mathcal N_k(\boldsymbol{\rho}) = \boldsymbol{ \rho} - \mathcal D_k (\boldsymbol{\rho}).
\end{equation} 

$\mathcal D_k(\boldsymbol{\rho})$ can be thought of as a multichannel CNN in the Fourier domain. The image domain input $\boldsymbol \rho$ is first transformed to k-space as $\boldsymbol{\widehat \rho}$, then passes through the k-space model, and, then transformed back to the image domain. Figure~\ref{subfig:cnn} shows the proposed M-layer CNN architecture. The overall k-space MoDL-MUSSELS network architecture is shown in Fig.~\ref{subfig:dwdcKsp}, which solves Eq.~\eqref{eq:modlKsp}.  Unlike IRLS-MUSSELS in Fig.~\ref{fig:mussels}, the parameters of this
network are not updated within the iterations and are learned from the exemplar data.

\subsection{ Hybrid MoDL-MUSSELS  Regularization}
\label{sec:hybrid}

\begin{figure}
  \centering 
  \def\svgwidth{.99\linewidth}
  \input{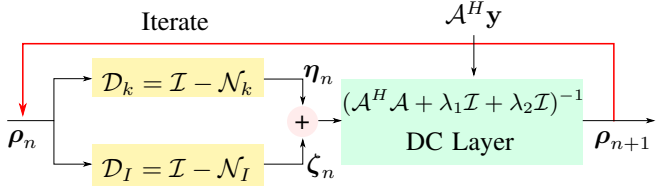}  
  \caption{The proposed hybrid MoDL-MUSSELS architecture
   resulting  from the alternating scheme shown in~\eqref{dcstep}-\eqref{secondshrinkage}. Here the $\mathcal D_k$ and $\mathcal D_I$ blocks represent the k-space and the   image-space denoising networks, respectively. The $\mathcal D_k$ and $\mathcal D_I$ networks have identical structures as shown in Fig~\ref{subfig:cnn}. The  learnable convolution weights are differnt for networks  $\mathcal D_k$ and $\mathcal D_I$ but remain constant across iterations.  }
  \label{fig:hybridModel}
\end{figure}

A key benefit of the MoDL framework over direct inversion methods is
the ability to exploit different kinds of priors, as shown in our prior work~\cite{modl-storm}.  The IRLS-MUSSELS and the
MoDL-MUSSELS schemes exploit the multichannel convolution relations
between the k-space data. By contrast, we relied on an image domain
convolutional neural network in earlier work~\cite{modl} to exploit the structure
of patches in the image domain. Note that this structure is completely
complementary to the multichannel convolution relations. We now
propose to jointly exploit both the priors as follows:
\begin{equation}
  \label{eq:modlKspImg}
  \argmin_{\PT} \|\AT(\boldsymbol \rho )-\vect y \|_2^2  + \lambda_1 \|\mathcal N_k(\PT)\|_2^2 + \lambda_2 \|\mathcal N_{I}( \boldsymbol \rho )\|_2^2,
\end{equation}
where $\mathcal N_k$ is the same prior as in~\eqref{eq:modlKsp}, while
$\mathcal N_I$ is an image space residual network of the form $\mathcal N_{I}( \boldsymbol \rho ) = \boldsymbol \rho - \mathcal D_{I}(\boldsymbol\rho )$.
Here, $\mathcal D_I$ is an image domain CNN as in earlier work~\cite{modl}. The problem \eqref{eq:modlKspImg} can be rewritten as \begin{equation*}
  \argmin_{\PT} \|\AT(\PT )-\vect y \|_2^2  + \lambda_1 \| \PT-\mathcal D_k(\PT)\|_2^2 + \lambda_2 \| \PT - \mathcal D_{I}(\PT )\|_2^2.
\end{equation*}
By substituting $\boldsymbol\eta=\mathcal D_k(\PT)$, and $\boldsymbol\zeta=\mathcal D_I(\PT)$,  an alternating minimization-based solution to the above problem iterates between the following steps:
\begin{align}  \label{dcstep}
  \boldsymbol\rho_{n+1}& =(\mathcal A^H \mathcal A + \lambda_1\mathcal I +\lambda_2 \mathcal I )^{-1} (\mathcal A^H\mathbf y + \lambda_1 \boldsymbol\eta + \lambda_2\boldsymbol\zeta)\\ \label{etasubproblem}
  {\boldsymbol{\eta}}_{n+1}&= \mathcal D_k({\boldsymbol{\rho}}_{n+1}) \\ \label{secondshrinkage}
  \boldsymbol{\zeta}_{n+1}&= \mathcal D_I ({\boldsymbol{\rho}_{n+1}}).
\end{align}

The above solution results in the hybrid MoDL-MUSSELS architecture shown in Fig.~\ref{fig:hybridModel}. Note that this alternating minimization scheme is similar to the plug-and-play priors\cite{chan2017plug} widely used in inverse problems. The main exception is
that we train the resulting network in an end-to-end fashion. Note
that, unlike the plug-and-plug denoisers that learn the image manifold,
the network $\mathcal D_k$ is designed to exploit the redundancies
between the multiple shots resulting from the annihilation relations. This
non-linear network is expected to project the multichannel k-space
data orthogonal to the nullspaces of the multichannel Hankel
matrices.
The regularization parameters $\lambda_1$ and $\lambda_2$ control the contribution of the k-space network and the image-domain network, respectively. During experiments we kept the values of $\lambda_1=0,01, \lambda_2=0.05$ fixed. However, it can be noted that these values can be made trainable as in the MoDL\cite{modl}.

\section{Experiments}
We perform several experiments to validate different aspects of the proposed model, such as, the benefits of the recursive network, the impact of regularization, robustness to  outliers, comparison with existing deep learning models such as U-NET\cite{ronneberger2015unet}, and comparison with a model-based technique  P-MUSE\cite{pocsmuse}.  
\subsection{Dataset Description}
\label{datadescription}

In vivo data were collected from healthy volunteers at the University of Iowa in accordance with the Institutional Review Board recommendations. The imaging was performed on a GE MR750W 3T scanner using a 32-channel head coil. A Stejskal-Tanner spin-echo diffusion imaging sequence was used with a 4-shot EPI readout. A total of $60$ diffusion gradient direction measurements were taken with a b-value of $700$~s/mm$^2$. The relevant imaging parameters were FOV~=~$210 \times 210$~mm, matrix size = $256\times 152$~ with partial Fourier oversampling of 24 lines, slice thickness~=~$4$~mm and TE = $84$~ms. Data were collected from 7 subjects.

The training dataset constituted a total of $68$~slices, each having $60$~directions and $4$~shots, from $5$~subjects. The validation was performed on $6$~slices of the $6$th subject, whereas testing was carried out on  $5$~slices of the $7$th~subject.  Thus, a total of $4080$,
$360$, and $300$ complex images each having size $256 \times 256 \times 4$ ($ \text{rows} \times \text{columns} \times \text{shots}$) were used for training, validation, and testing, respectively.

To perform quantitative comparisons, we also made use of simulated data with high SNR. For this purpose, we utilized a subset of pre-processed, relatively high-SNR diffusion dataset from the human connectome project~\cite{hcpDiffusion}. We extracted 15 volumes and 20 slices from 100 subjects, which resulted in 30,000 magnitude images of size $145 \times 174$. We prepared a dataset of 23,000 training images, 3,000 validation images, and 3,000 test images.We simulated the sensitivity maps using Walsh algorithm~\cite{walshcsm}.  To simulate the multishot data with phase errors, we multiplied each magnitude image with  synthetically generated random bandlimited phase errors using the image formation  model in Eq.~\eqref{modifiedforwardmodel}. Gaussian noise of varying amounts of standard deviation $\sigma$ was added to the phase-corrupted k-space data. The k-space data was under-sampled to generate the multishot data.

\subsection{Multi-coil forward model}
All of the model-based schemes used in this study (MUSE, MUSSELS, MoDL-MUSSELS) rely on a forward model that mimics the image formation. We implement this forward model as described in \eqref{fwd} and \eqref{modifiedforwardmodel}. The raw dataset consists of 32 channels. We reduce the data to  four virtual channels using singular value decomposition~(SVD) of the non-diffusion weighted (b0) image. The coil sensitivity maps of these four virtual channels were estimated using ESPIRIT \cite{espirit2014}. The same channel combination weights were used to reduce the diffusion-weighted MRI data to four coils.

\subsection{Quantitative metrics used in experiments}
The reconstruction quality is measured using the structure similarity index~(SSIM)~\cite{ssim} and peak signal-to-noise ratio~(PSNR). The PSNR is defined as
\begin{equation*}
  \mathrm{PSNR}(\mathbf x,\mathbf y)= 10*\log_{10}\left( \frac{\max(\mathbf x)^2}{\mathrm{MSE}(\mathbf x,\mathbf y)}   \right)
\end{equation*}
where MSE is the mean-square-error between $\mathbf x$ and $\mathbf y$. The final PSNR/SSIM value is estimated by the average of the PSNR/SSIM of individual shots.

\subsection{Algorithms used for comparison}

We compare the proposed scheme against IRLS-MUSSELS~\cite{irlsMussels},P-MUSE~\cite{pocsmuse}, and a solution based on U-NET~\cite{ronneberger2015unet}. The IRLS-MUSSELS is a modification of the  MUSSELS algorithm~\cite{mussels}. Specifically, the modification involve an IRLS based implementation instead of iterative shirnkage algorithm in~\cite{mussels}, which results in a faster implementation. Moreover, it also includes an additional conjugate symmetry constraint in addition to the annihilation relations between the shots that is exploited in the  MUSSELS method~\cite{mussels}. We refer the readers to~\cite{irlsMussels}, which shows that the addition of the conjugate symmetry constraint reduces blurring and results in sharper images compared to the original MUSSELS method~\cite{mussels} for partial Fourier acquisitions. In the results section,  the IRLS-MUSSELS is referred to  as simply IRLS-M.

P-MUSE~\cite{pocsmuse} is a two-step algorithm that first estimates the motion-induced phase using the SENSE\cite{sense1999} reconstruction and the total-variation denoising. With the knowledge of the phase errors, it recovers the images using a regularized optimization with \eqref{modifiedforwardmodel} as the forward model.  The P-MUSE algorithm\cite{pocsmuse} has three parameters $\lambda_1=0.01$, $\lambda_2=0.01$, and the number of iterations~=~40. The parameters $\lambda_1$ and $\lambda_2$ control the total variation regularization during phase estimation and reconstruction, respectively.  We searched over the parameters to yield the best possible reconstruction.

 We extended the U-NET~\cite{ronneberger2015unet} model for the multishot diffusion MR image reconstruction. The number of convolution layers, the feature maps in each layer, and the filter size were kept the same as in~\cite{ronneberger2015unet}. The input to the extended U-NET model was the  concatenation of  the real and  imaginary parts of phase-corrupted coil-combined complex 4 shots images. The IRLS-MUSSELS~\cite{irlsMussels} reconstructions were used as the ground truth for  the training of the deep learning models on experimental data. We trained the network in the image domain with $1,000$ epochs for $13$ hours using the Adam\cite{Kingma2015} optimizer.

We  also performed a comparison between the k-space MoDL-MUSSELS formulation in section~\ref{sec:kspace}  and the hybrid MoDL-MUSSELS formulation in section~\ref{sec:hybrid}. We refer to the former as the k-space network and the latter as the hybrid network.
To perform a fair comparison between hybrid and k-space networks, the number of parameters in the k-space  network was kept the same as that of the hybrid model by increasing the number of feature maps in the convolution layers.

\subsection{Network architecture and training }
\begin{figure}
  \centering
\def\svgwidth{.85\linewidth}
\input{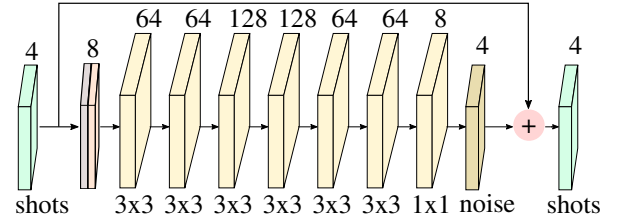}
\caption{The specific M=8 layer residual learning CNN architecture used as $\mathcal D_k$ and $\mathcal D_I$ blocks in the  experiments. The 4 shot complex data are the input and output of the network. The first layer concatenates the real and imaginary parts as  8 input features. The numbers on top of each layer represent the number of feature maps learned at that layer. We learn $3\times 3$ filters at each layer except the last, where we learn $1\times 1$ filter. }
\label{fig:nn7lay}
\end{figure}

In this work, we trained an $8$-layer CNN having  convolution filters  of size $3\times 3$  in each layer. Each layer comprises a convolution, followed by ReLU, except the last layer, which consists of a $1 \times 1$ convolution as shown in Fig.~\ref{fig:nn7lay}. The  real and imaginary components of the complex  4 shots  data were considered as channels in the residual learning CNN architecure, whereas the data-consistency block worked explicitly with complex data.

\begin{figure}
  \centering \begin{tikzpicture}
 \begin{axis}[
   width=.9\linewidth,
   height=2.2in,
   axis y line*=left,
   xlabel=Number of training epochs,
   ylabel=Training loss,
   grid=both,   
   grid style={line width=.1pt, draw=gray!10},
   major grid style={line width=.2pt,draw=gray!50},
   minor tick num=1,
   legend style={at={(.87,0.95)},draw=none},
   ]
   \addplot[mark=.,blue,line width=1] table [x=epoch, y=tloss,col sep=comma] {loss.csv};
   \addlegendentry[text width=2cm]{Training Loss}
 \end{axis}
 
   \begin{axis}[
   width=.9\linewidth,
   height=2.2in,
     axis y line*=right,
     axis x line=none,
     ylabel=Validation loss,
   legend style={at={(.915,0.8)},draw=none},
   ]
   \addplot[mark=.,red, line width=1] table [x=epoch, y=vloss, col sep=comma] {loss.csv};
    \addlegendentry[text width=2.3cm]{Validation Loss}
    \end{axis}

    
\end{tikzpicture}

  \caption{ The decay of training and validation errors with
    epochs. Each epoch represents one sweep through the entire dataset. We note that both the losses decay with iterations. This suggests that the amount of training data is sufficient to train the parameters of the model.
  }
  \label{plt:loss}
\end{figure}
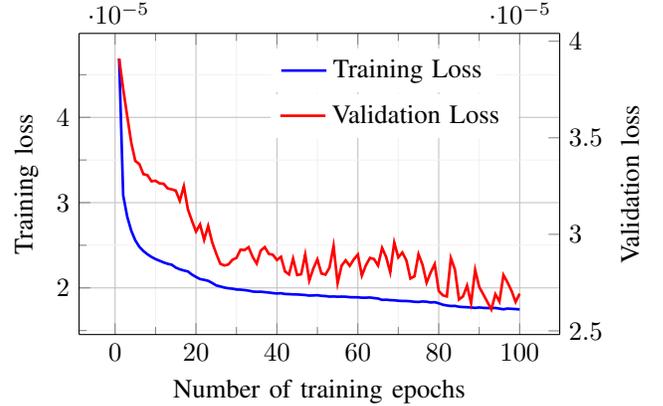

The proposed network architecture, as shown in Fig.~\ref{fig:hybridModel}, was  unfolded for three iterations,  and the end-to-end training  was performed for $100$~epochs. The input to the unfolded network is the zero-filled complex data from the four shots, which corresponds to $A^H \mathbf y$, while the network outputs the fully sampled complex data for the four shots. The proposed MoDL-MUSSELS architecture combines the data from the four shots using the sum-of-squares approach. The network weights were randomly initialized using Xavier initialization and shared between iterations. The network was implemented using the TensorFlow library in Python 3.6 and trained using the NVIDIA P100 GPU. The conjugate-gradient optimization in the DC step was implemented as a layer operation in the TensorFlow library as described in the work~\cite{modl}. We utilized the mean-square-error as the loss function during training. The total network training time of the network was around 37 hours.

The plot in Fig.~\ref{plt:loss} shows training loss decays smoothly with epochs. It can be noted that the loss  on the validation dataset also has overall decaying behavior, which implies that the trained model did not over-fit the dataset. The model-based framework has considerably fewer parameters than direct inverse methods and hence requires far fewer training data to achieve good performance, as seen from the  experiments in the previous work~\cite{modl}.

\section{Results}

\subsection{Comparisons using simulated data}

\begin{table} 
\caption{The PSNR (dB) and SSIM values obtained by five methods on the testing dataset with simulated phases  and added Gaussian noise of varying standard deviation $\sigma$. The values are reported as mean~$\pm$~standard deviation.}
\label{tab:simulation}
\centering
\begin{tabular}{lccc} \toprule 
              & \multicolumn{3}{c}{Peak signal to noise ratio (dB)}  \\ \midrule
Noise (std)   & $\sigma=0.001$  & $\sigma=0.002$  & $\sigma=0.003$   \\ \midrule
U-NET         & $32.15 \pm 2.12$  & $29.98 \pm 1.19$ & $27.63 \pm 0.82$     \\
P-MUSE          & $34.08 \pm 2.31$  &$ 31.68 \pm 2.21$ & $29.19 \pm 1.84$  \\
IRLS-M       & $38.81 \pm 1.98$  &$ 36.21 \pm 1.32$ & $32.43 \pm 1.33$ \\
K-space        & $40.02 \pm 1.18$  & $36.92 \pm 0.96$ & $34.69 \pm 1.38$     \\
Hybrid        & $40.59 \pm 1.87$  &$ 37.37 \pm 1.56$ & $35.40 \pm 1.36$ \\ \midrule
              & \multicolumn{3}{c}{Structural similarity index}  \\ \midrule
U-NET         &$0.89 \pm 0.01$ & $ 0.82 \pm 0.02$  & $0.73 \pm 0.03$    \\
P-MUSE          &$0.79 \pm 0.03$ & $ 0.69 \pm 0.04$  & $0.63 \pm 0.05$    \\
IRLS-M       &$0.88 \pm 0.01$ & $ 0.83 \pm 0.01$  & $0.72 \pm 0.03$    \\
K-space        &$0.94 \pm 0.01$ & $ 0.89 \pm 0.02$  & $0.84 \pm 0.03$    \\
Hybrid        &$0.96 \pm 0.00$ & $ 0.94 \pm 0.01$  & $0.92 \pm 0.01$    \\  \bottomrule
\end{tabular}
\end{table}

Table~\ref{tab:simulation} summarizes the quantitative results (PSNR and SSIM values) obtained from the simulated data in Section~\ref{datadescription}. Specifically, we quantitatively compare the reconstructions provided by the five algorithms, while varying the noise levels.
We did not perform the training of the deep learning methods (k-space, U-NET, and hybrid) for different noise levels but instead utilized the same model trained for a single noise level ($\sigma=0.001$).
We adjusted the parameters of the P-MUSE and IRLS-MUSSELS algorithms for different noise levels to get the best average results. The average performance of the U-NET is lower than all other methods since the U-NET does not have an explicit data-consistency term like the other methods. It is evident from the graphs in  Fig.~\ref{fig:ssim_plot} that proposed hybrid method performs better than other methods on all individual slices and directions of a test subject.

\begin{figure} \centering
  \includegraphics[width=.8\linewidth]{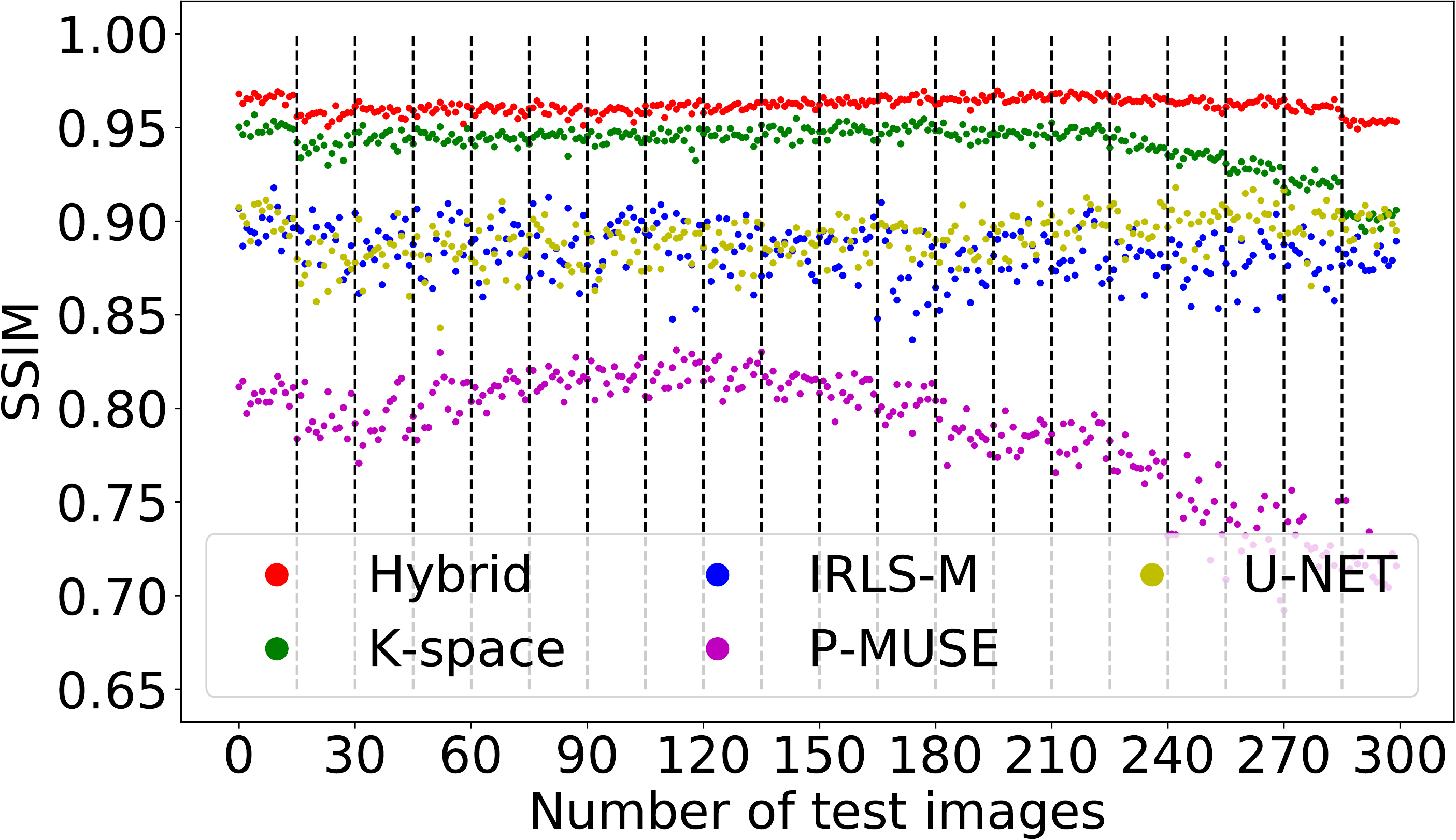}
  \caption{ This plot compares the variation of the SSIM on all the slices and directions of one test subject from
simulation dataset. The vertical lines seperate the different slices, i.e., first fifteen images are the directions
corresponding to the first slice, and so on. There is a total of 20 slices, each having 15 different directions,
resulting in a total of 300 images.}
\label{fig:ssim_plot}
\end{figure}

\begin{figure*}
\input{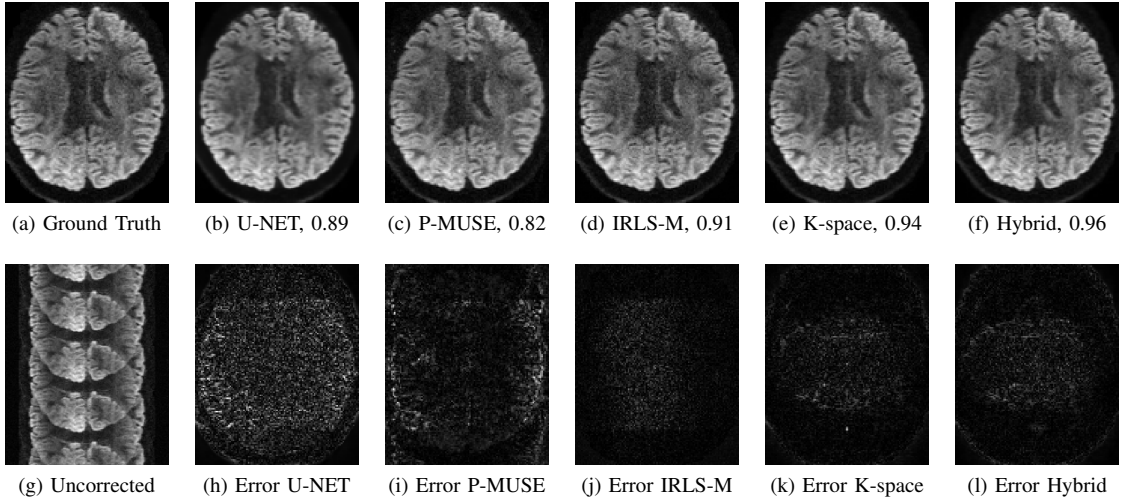}
\caption{Simulation results using the high-SNR data obtained from HCP. The reconstructed images and the corresponding error maps from five different algorithms are shown. Here, the ground truth  is an image from the test dataset that was corrupted with phase errors of bandwidth 3x3 and noise standard deviation $\sigma=0.001$ to simulate 4 shot acquisition. (g) shows the uncorrected image if we do not correct the phase errors during reconstruction. The numbers in the sub-captions represent the SSIM values.  }
\label{fig:simulation}
\end{figure*}

Figure~\ref{fig:simulation} shows an example set of images reconstructed using the five methods for this simulated data. For comparison, the uncorrected image and the error maps for all the reconstructions, compared to the ground truth image are also provided.  It is evident from the error maps that the proposed hybrid model has the least error among  the methods compared. Figure~\ref{fig:simulation}(e) shows that the  k-space network  is able to compensate for phase errors of multishot data.  The  addition of image-domain regularization in the hybrid model further improves the reconstructions in Fig.~\ref{fig:simulation}(f). We note that the image
domain network exploits the manifold structure of patches, which
serves as a strong prior that the k-space network has difficulty
capturing.

\subsection{Robustness to outliers}
\begin{figure}
  \input{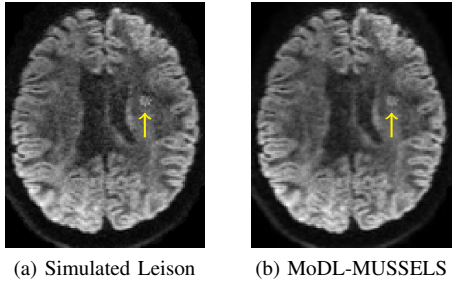}
  \caption{Lesion experiment. Arrow points to the location of Lesion. The proposed MoDL-MUSSELS method  preserves the lesion.}
  \label{fig:lesion}
\end{figure}

We further performed an experiment to determine the robustness of the proposed MoDL-MUSSELS approach against outliers.
In particular, we simulated a lesion image by increasing the intensity at a few pixels as indicated by an arrow in Fig.~\ref{fig:lesion}(a).  This image was passed through  the existing trained model. It is observed from  Fig.~\ref{fig:lesion}(b) that simulated lesion was preserved by the proposed method.  Note that the training dataset did not have any lesion images and we did not simulate the lesion images during training.
The robustness of the algorithm to such outliers can be attributed to the fact that the algorithm relies on k-space and q-space deep-learning networks with small receptive fields, unlike direct inversions methods that rely on large receptive fields. Hence, the proposed scheme learns only to exploit local redundancies in k-space and q-space and does not memorize whole images.

\subsection{Comparison  of reconstruction time}

\begin{table}
  \caption{ Time to reconstruct all five slices of the test
    subject. Each slice had 60 directions, 4 shots, and 32 coils.  IRLS-MUSSELS and P-MUSE  were runs on CPU with parallel processing.  }
  \label{tab:time} \centering
  \begin{tabular}{ccccc} \toprule Algorithm: & U-NET & P-MUSE & IRLS-M
    & MoDL-MUSSELS \\ \midrule Time (sec) : &7 & 632& 1386 & 49 \\
    \bottomrule
  \end{tabular}
\end{table}

Table~\ref{tab:time} compares the time taken to reconstruct the entire testing dataset for  various methods.
It is noted that the computational complexity of the MoDL-MUSSELS is around 28-fold lower than the IRLS-MUSSELS. Note that IRLS-MUSSELS estimates the optimal linear filter bank from the measurements itself, which requires significantly many iterations. By contrast, MoDL-MUSSELS pre-learns  non-linear network weights.  The quite significant speed increase  directly follows from the significantly fewer number of iterations. Note that we rely on a conjugate-gradient algorithm to enforce data consistency specified by~\eqref{dcstep}. Also note that solving \eqref{dcstep} exactly as opposed to the use of steepest gradient steps at each iteration would require
more  unrolling steps, thus diminishing the gain in speedup. The greatly reduced runtime is expected to facilitate the
deployment of the proposed algorithm on clinical scanners.

\subsection{Impact of iterations on image quality}

\begin{figure}
  \input{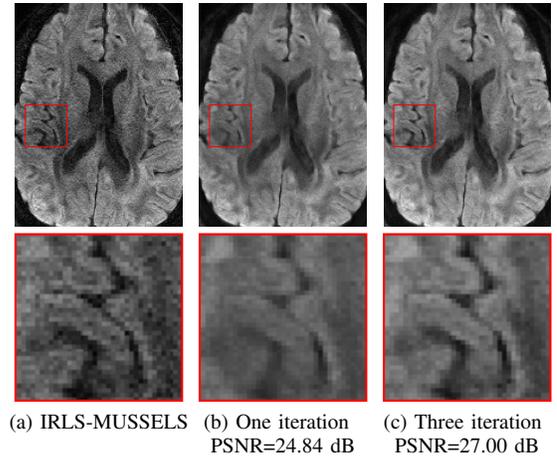}
              
  \caption{Effect of iterations on image quality. We observe that the quality of the reconstructions with the proposed MoDL-MUSSELS scheme improve with iterations. Specifically, the sharpness of the image and the contrast seem to improve with more iterations.  }
  \label{fig:iterationEffect}
\end{figure}

 Figure~\ref{fig:iterationEffect} shows the impact of the number of iterations in the iterative algorithm described in
\eqref{dcstep}-\eqref{secondshrinkage}.  Specifically, we unrolled the iterative
algorithm for the different numbers of iterations and compared the performance of the resulting networks. We used the hybrid model due to its improved performance. The parameters of
both the k-space and image-space networks were shared across iterations.  Specifically, MoDL-MUSSELS
uses three iterations of alternating-minimization, with five iterations of CG within each alternating step. The IRLS-MUSSELS uses five iterations of both outer loop as well as CG step.
The images in Fig.~\ref{fig:iterationEffect} each correspond to a specific direction and slice in the testing dataset. We note that the contrast and details in the image improved with iterations, as did the visualization of some features, as shown in the zoomed portions.

\subsection{Comparisons on experimental data}

\begin{figure}
  \input{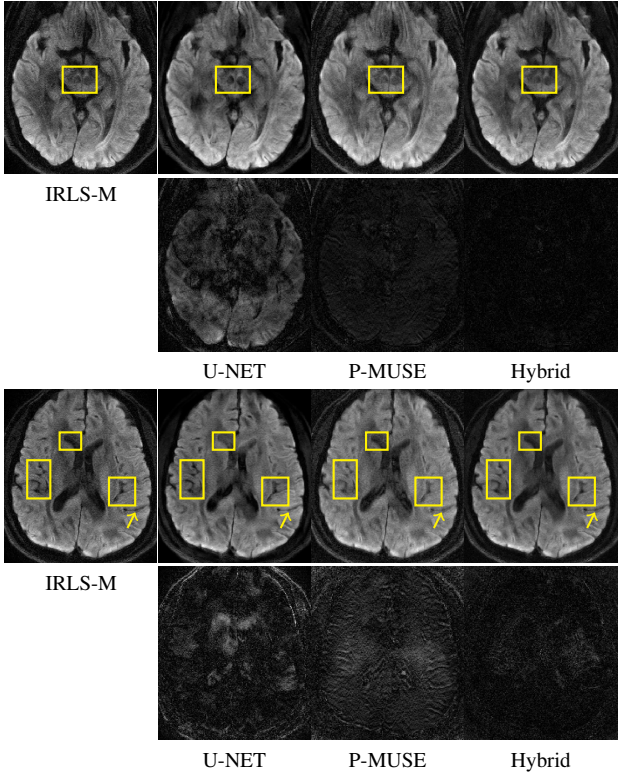}
  \caption{Reconstructions obtained using different algorithms on experimental partial-Fourier data. Row 1 and Row 3 shows reconstruction results from two different slices.  We generated the error maps in rows~2~and~4 by considering MUSSELS reconstructions as ground truth.  The yellow boxes highlight the differences.}
  \label{fig:compare}
\end{figure}

Next we compare the performance of the proposed method to reconstruct experimental data.
Figure~\ref{fig:compare} shows the reconstructions offered by the different algorithms. A separate network was trained with the experimental data utilizing the IRLS-MUSSELS as the ground truth.
While this comparison may not be fair to P-MUSE, we used this approach since the main goal is to validate the MoDL-MUSSELS and U-NET which relied on IRLS-MUSSELS results for training.
As evident from the error maps, the U-NET reconstructions appear less blurred, but it seems to miss some key features highlighted by yellow boxes.
The hybrid method provides good results comparable to that of IRLS-MUSSELS.

To further validate the reconstruction accuracy of all the DWIs corresponding to the test slice, we performed a tensor fitting using
all the DWIs and compared the resulting fractional anisotropy (FA) maps and the fiber orientation maps. For this purpose, the DWIs
reconstructed using various methods were fed to a tensor fitting routine (FDT Toolbox, FSL).  The FA maps were computed
from the fitted tensors, and the direction of the primary eigenvectors of the tensors was used to estimate the fiber orientation. The FA maps
generated using the various reconstruction methods are shown in Fig.~\ref{fig:fa2}, which has been color-coded based on the fiber
direction. It is noted that these fiber directions reconstructed by the IRLS-MUSSELS method and the MoDL-MUSSELS match the true anatomy known
for this brain region from a diffusion tensor imaging~(DTI) white matter atlas~(\url{http://www.dtiatlas.org}).

\begin{figure}
  \centering
  \includegraphics[width=.99\linewidth]{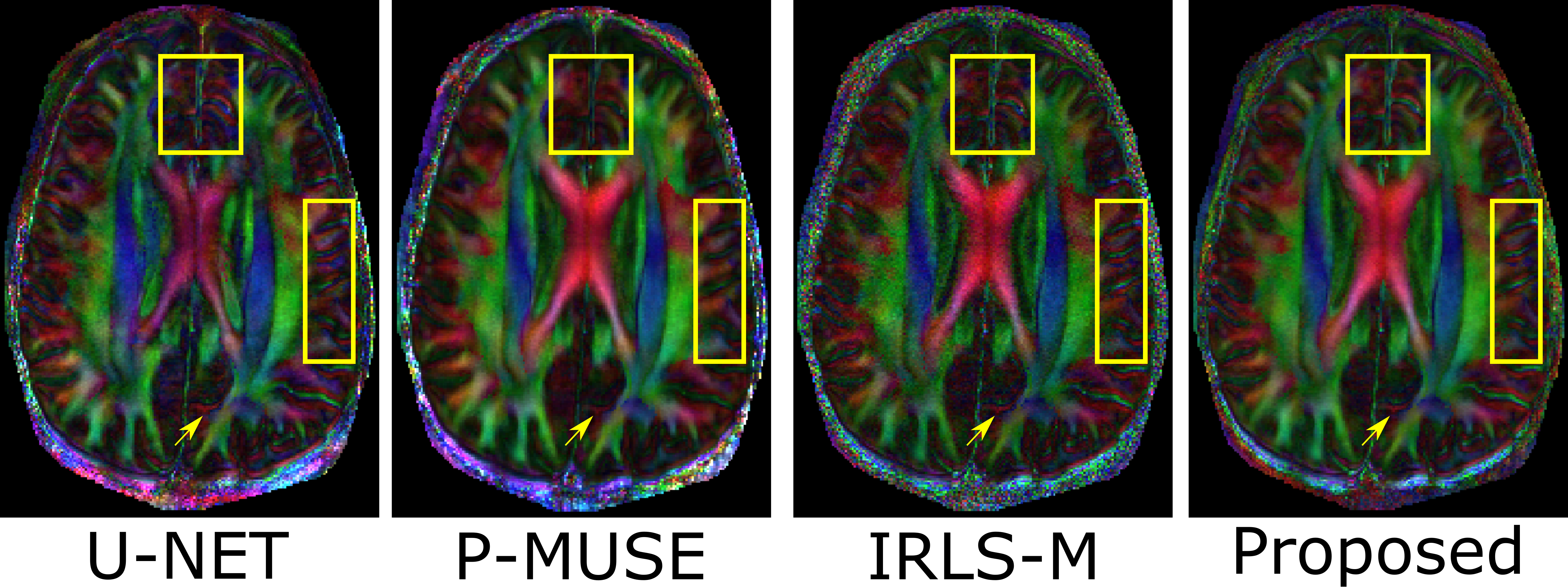}
  \caption{The fractional-anisotropy maps on a test dataset slice. These images are computed from the sixty directions of the slices, recovered using the respective algorithms. We note  the proposed scheme provides less blurred reconstructions than P-MUSE, which are comparable with IRLS-MUSSELS.}
  \label{fig:fa2}
\end{figure}

\section{Discussion}

We observe that the IRLS-MUSSELS reconstructions on experimental data are  noisy. This noisy ground truth training data causes fuzziness in the training loss, which translates to the slight blurring in  MoDL-MUSSELS reconstructions in  Fig.~\ref{fig:compare}. Note that the MoDL-MUSSELS reconstructions in the simulated data experiments in Fig.~\ref{fig:simulation} are  less blurred. The dependence of the final image quality on the training data is a limitation of the current work, especially in the multishot diffusion setting where noise-free training data is difficult to acquire. We plan to experiment with denoising strategies as well as the acquisition of training data with multiple averages to mitigate these problems.
Further, we note from Fig.~\ref{fig:compare} that the reconstructions provided by the MoDL-MUSSELS appear less noisy and are visually more appealing than the noisy ground truth obtained using  the IRLS-MUSSELS.  This behavior may be attributed to the convolutional structure of the network, which is known to offer implicit regularization~\cite{deepImagePrior}.

In this work, we utilized an eight-layer neural network, as shown in  Fig.~\ref{fig:nn7lay}. However, the proposed MoDL-MUSSELS architecture in  Fig.~\ref{fig:hybridModel}  is not constrained by  choice of  network. Any network architecture (e.g., U-NET) may be used instead. It is possible that the results can improve by utilizing more sophisticated network architecture. Further, it can be noted that the proposed model architecture is flexible to allow different network architectures for image-space and k-space models. However, for the proof of concept, we used the same network architecture for both
k-space and image space.

To avoid overfitting the model and reduce the training time, the proposed network in Fig.~\ref{fig:hybridModel} was unfolded for three iterations before performing the joint training. The sharing of network parameters allows the network to be unfolded for any number of iterations without increasing the number of trainable parameters. In this work, we restricted our implementation to a three iteration setting. We note that the results may improve with more outer iterations. However, increasing the outer iterations require more GPU memory during network training.

The deep learning blocks used in the proposed scheme map the noisy/artefact-prone N-shot data to the noise-free N-shot data.  The size of the filters in the first and last layers of the deep learning blocks depend on the number of shots. Hence, the network needs to be retrained if the number of shots changes. Since the filters capture the annihilation relations between the shots, we do not anticipate the need to retrain the network if other parameters (e.g., image size, TR, TE, etc.) change. Finally, we note that the current method depends on the estimation of the coil sensitivities to recover the multishot data.

\section{Conclusions} 
We introduced a model-based deep learning framework termed MoDL-MUSSELS for the compensation of phase errors in multishot diffusion-weighted MRI data. The proposed algorithm  alternates between a conjugate gradient optimization algorithm to enforce data
consistency and multichannel convolutional neural networks (CNN) to project the data to appropriate subspaces. We rely on a hybrid approach involving a multichannel CNN in the k-space and another one in the image space. The k-space CNN exploits the annihilation relations between the shot images, while the image domain network is used to project the data to an image manifold. The weights of the deep network, obtained by unrolling the iterations in the iterative optimization scheme, are learned from exemplary data in an end-to-end fashion. The experiments show that the proposed scheme can yield reconstructions that are comparable to state-of-the-art methods while offering several orders of magnitude reduction in run-time. 

\section{Acknowledgment}
Data were provided [in part] by the Human Connectome Project, WU-Minn Consortium (Principal Investigators: David Van Essen and Kamil Ugurbil; 1U54MH091657) funded by the 16 NIH Institutes and Centers that support the NIH Blueprint for Neuroscience Research; and by the McDonnell Center for Systems Neuroscience at Washington University.

\bibliographystyle{IEEEtran}


\end{document}